%% file: main.tex
\documentclass[10pt,twocolumn,letterpaper]{article}

\usepackage[pagenumbers]{cvpr} 

\usepackage[accsupp]{axessibility}  
\usepackage{graphicx}
\usepackage{amsmath}
\usepackage{amssymb}
\usepackage{booktabs}
\usepackage{dsfont}

\usepackage[table]{xcolor}
\definecolor{LightGreen}{rgb}{0.93,0.98,0.96}

\newcommand{\datasetname}{\textit{This-Is-My}\xspace}

\newcommand{\possesivepatterns}{\textbf{List of possessive text patterns:} \texttt{<this is my>}, \texttt{<this is our>}, \texttt{<this is his>}, \texttt{<this is her>}, \texttt{<this is their>}, \texttt{<these are my>}, \texttt{<these are our>}, \texttt{<these are his>}, \texttt{<these are her>}, \texttt{<these are their>}}

\usepackage[pagebackref,breaklinks,colorlinks]{hyperref}

\usepackage[capitalize]{cleveref}
\crefname{section}{Sec.}{Secs.}
\Crefname{section}{Section}{Sections}
\Crefname{table}{Table}{Tables}
\crefname{table}{Tab.}{Tabs.}

\DeclareMathOperator*{\argmin}{arg\,min}

\def\cvprPaperID{5983} 
\def\confName{CVPR}
\def\confYear{2023}

\begin{document}

\title{Meta-Personalizing Vision-Language Models to Find Named Instances in Video}

\author{Chun-Hsiao Yeh\textsuperscript{1,3*} \quad 
Bryan Russell\textsuperscript{3} \quad 
Josef Sivic\textsuperscript{2,3} \quad 
Fabian Caba Heilbron\textsuperscript{3}{$^\dagger$} \quad 
Simon Jenni\textsuperscript{3}{$^\dagger$} \vspace{4pt}\\
\textsuperscript{1}University of California, Berkeley \qquad
\textsuperscript{2}CIIRC CTU \qquad
\textsuperscript{3}Adobe Research   \\
{\tt\small daniel\_yeh@berkeley.edu \quad \{brussell,inr03127,caba,jenni\}@adobe.com}   \\
}

\maketitle
{\let\thefootnote\relax\footnote{{$^\dagger$}Equal advising.}}

{\let\thefootnote\relax\footnote{$^*$Work done during CHY's summer internship at Adobe Research.}}

{\let\thefootnote\relax\footnote{$^2$Czech Institute of Informatics, Robotics and Cybernetics at the Czech Technical University in Prague.}}

\input{sections/01_abstract.tex}

\input{sections/02_introduction.tex}

\input{sections/03_related.tex}

\input{sections/04_method.tex}

\input{sections/05_dataset.tex}

\input{sections/06_results.tex}

\input{sections/07_conclusion.tex}

{\small
\bibliographystyle{ieee_fullname}
\bibliography{egbib}
}

\input{sections/appendix}

\end{document}

%% file: sections/01_abstract.tex
\begin{abstract}
Large-scale vision-language models (VLM) have shown impressive results for language-guided search applications. 
While these models allow category-level queries, they currently struggle with personalized searches for moments in a video where a specific object instance such as ``My dog Biscuit'' appears. 
We present the following three contributions to address this problem.
First, we describe a method to {\em meta-personalize} a pre-trained VLM, \ie, learning how to learn to personalize a VLM at test time to search in video. 
Our method extends the VLM's token vocabulary by learning novel word embeddings specific to each instance. 
To capture only instance-specific features, we represent each instance embedding as a combination of shared and learned global category features.
Second, we propose to learn such personalization without explicit human supervision.
Our approach automatically identifies moments of named visual instances in video using transcripts and vision-language similarity in the VLM's embedding space.
Finally, we introduce \datasetname, a personal video instance retrieval benchmark. We evaluate our approach on \datasetname and DeepFashion2 and show that we obtain a 15\% relative improvement over the state of the art on the latter dataset.

\end{abstract}

%% file: sections/02_introduction.tex
\section{Introduction}
\label{sec:intro}

\begin{figure}[t]
    \centering
    \includegraphics[width=\linewidth]{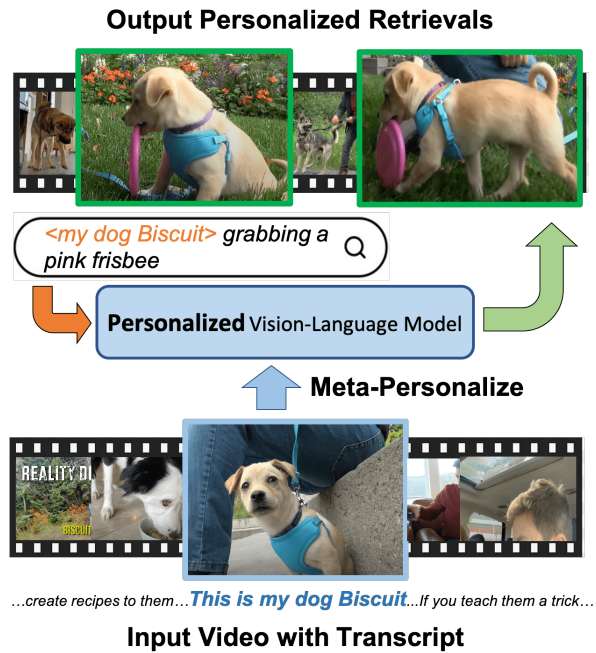}
    \vspace{-7mm}
    \caption{\textbf{Meta-Personalized Vision-Language Model (VLM) to Retrieve Named Instances in Video.}
    Given a video where a user-specific instance, \eg, ``My dog Biscuit'' is mentioned, our method automatically learns a representation for the user-specific instance in the VLM's text input space.
    The personalized VLM can then be used to retrieve the learned instance in other contexts through natural language queries, \eg, \texttt{$<$my dog Biscuit$>$ grabbing a pink frisbee}. This result is enabled by  meta-personalizing the VLM on a large-scale dataset of narrated videos by pre-learning shared global category tokens (in this example for the category of 'dogs'), which are then easily personalized to user-specific instances from only a few user-given training examples.  
    }
    \vspace{-3mm}
    \label{fig:teaser}
\end{figure}

The recent introduction of large-scale pre-trained vision-language models (VLMs) has enabled many new vision tasks, including zero-shot classification and retrieval~\cite{radford2021learning,li2021align,jia2021scaling}, image/video generation~\cite{Ramesh2021ZeroShotTG, Ramesh2022HierarchicalTI, Saharia2022PhotorealisticTD, Singer2022MakeAVideoTG, Ho2022ImagenVH, Villegas2022PhenakiVL}, or language-guided question answering~\cite{Alayrac2022FlamingoAV,li2022composing, yang2022zero}.
It is now possible to search not only for specific object categories (\eg, dogs) but also for more specific descriptions of both the object and scene attributes (\eg, ``A small white dog playing at the dog park'').  
However, we often do not want to search for just any example of a generic category but instead to find a specific instance. 
For example, a user might want to search their personal video library for all the scenes that show their dog ``Biscuit grabbing a pink frisbee'', as illustrated in Figure~\ref{fig:teaser}. 
Since VLMs do not have a representation of ``Biscuit,'' such queries are beyond the capabilities of off-the-shelf VLMs. 

Recent work \cite{cohen2022my} proposed a method to extend the language encoder's vocabulary with a newly learned token that represents a specific personal instance to address this issue. 
While this approach enables language-guided search for personal instances by placing the learned tokens in the query prompt, their solution assumes a collection of manually annotated images showing the individual instance in various contexts for successful token learning. 
For this approach to work in practice, a user must manually annotate all their important personal instances in various contexts, such that the instance representation does not capture nuisance features, \eg, the background. 
We thus identify two key challenges: 1) collecting personal instance examples without explicit human labeling and 2) learning a generalizable object-centric representation of personal instances from very few examples. 

The contributions of this work are three-fold. As our first contribution, we propose a method to automatically identify important personal instances in videos for personalizing a vision-language model without explicit human annotations. 
Indeed, people often record and refer to personal items or relationships in videos found online.
Our approach is thus to identify mentions of personal instances in a video automatically and leverage these moments to build a set of personal instances for training. 
To this end, we extract the transcripts of videos using speech-to-text models and find candidate moments by looking for occurrences of ``this is my *'' or similar possessive adjective patterns.
The symbol * in this example could represent a single word or sequence of words describing the instance (\eg, *= ``dog Biscuit''). 
We then use vision-language similarity to filter non-visual examples and to find additional occurrences in the video for training.
For example, we found more than six thousand named instances in 50K videos randomly sampled from the Merlot Reserve dataset \cite{zellers2022merlot}.
We call the resulting collection of named instances in videos the \datasetname dataset. 

As our second contribution, we propose a novel model and training procedure to learn text tokens representing the named instances in video from possibly very few and noisy training examples. 
Our method represents each instance with learned tokens and models each token as a linear combination of a set of pre-learned category-specific features shared across different instances.
This set of shared category-specific features (similar to object attributes) improves the generalization of our method by preventing the instance representations from capturing nuisance features (\eg, the scene background). 
Furthermore, we show how to pre-train and adapt the shared category features using a large set of automatically collected \datasetname examples, further improving our model's few-shot personalization performance at test-time.  
We call this pre-training of shared category features {\em meta-personalization}. 
In contrast to prior work \cite{cohen2022my}, our method does not require training additional neural network models and requires only the optimization of a contrastive learning objective.

As our final contribution, we demonstrate and evaluate our model on an existing fashion item retrieval benchmark, DeepFashion2, and our new challenging \datasetname video instance retrieval dataset\footnote{Available at \url{https://danielchyeh.github.io/metaper/}} depicting specific object instances across different videos and contexts.
Our experiments demonstrate that our method outperforms several baselines and prior approaches on these challenging language-guided instance retrieval tasks. 

\begin{figure*}[t!]
\begin{center}
\includegraphics[width=\textwidth]{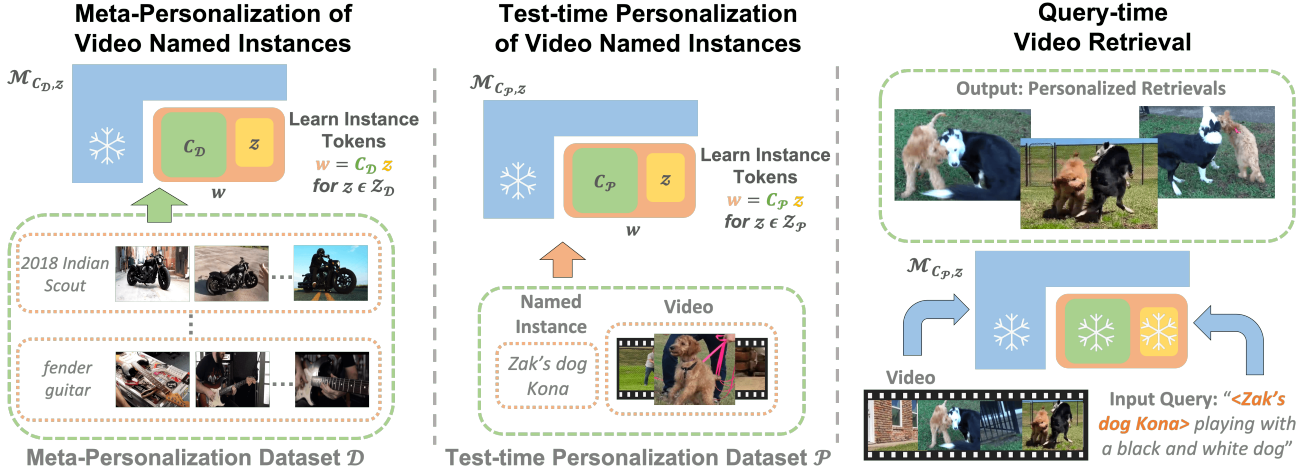}
\vspace{-8mm}
\caption{\textbf{Overview of our Personalized Vision-Language Model.} 
Our model augments a frozen VLM (blue) with novel personal instance tokens $w=C z$ (orange) that are a combination of global category features $C$ (green) with instance-specific weights $z \in \mathcal{Z}$ (yellow).
Our approach for personalized instance retrieval has three stages. First, we pre-learn global category features $C_\mathcal{D}$ on a large set of automatically mined named personal instances in videos. 
We call this process Meta-Personalization (left). 
In the second step (middle), we adapt the meta-personalized category features $C_\mathcal{D}$ at test-time and learn novel instance weights $z \in \mathcal{Z}_\mathcal{P}$ to represent a user's personal instances via $w=C_\mathcal{P} z$.
Finally (right), we leverage the (frozen) personalized instance tokens $w$ in natural language queries at query time.  
}
\vspace{-5mm}
\label{fig:overview}
\end{center}
\end{figure*}

%% file: sections/03_related.tex
\section{Related Work}
\label{sec:relatedwork}

\noindent \textbf{Vision-Language Models for Video Retrieval.} Vision-language foundational models~\cite{radford2021learning, jia2021scaling, li2021align} have been successful for zero-shot and other diverse video tasks, such as video question answering~\cite{li2022composing, yang2022zero}, language-video grounding~\cite{Huang2018FindingW, yang2022tubedetr, woo2022explore}, and text-to-video retrieval~\cite{xue2022clip, kunitsyn2022mdmmt, ma2022x, gorti2022x, luo2022clip4clip, xu2021videoclip, lei2021less}. These models have a powerful representation that transfers well to the video domain to achieve competitive performance on video-language tasks. Our approach builds on these powerful representations to retrieve specific named instances in video.

\noindent \textbf{Personalized Concept Learning.}
Adapting a model to learn a user-specific representation has been a significant topic in machine learning research, including recommendation systems~\cite{amat2018artwork,benhamdi2017personalized} and federated learning~\cite{jiang2019improving}. Relevant to us are recent approaches for adapting vision-language models to object instances. PALAVRA~\cite{cohen2022my} proposes a learning scheme that appends a learnable token for a new personalized concept to the token embedding of the input text prompt. This learned representation helps to preserve the personalized concept. DualPrompt~\cite{wang2022dualprompt} introduces a framework that learns a small set of prompts to emphasize more specific concepts without forgetting the learned concepts in the pre-trained foundational model.
There have also been works that have extended this personalized concept learning to image generation~\cite{casanova2021instance,Gal2022Inversion,ruiz2022dreambooth}. 
While these approaches adapt a vision-language model to personal instances, they perform the adaptation independently for each instance and do not ``meta-train'' for the personalization task for improved few-shot learning as we do in this work.

\noindent \textbf{Fine-tuning and Test-time Adaptation.}
Fine-tuning is a common strategy to adapt a pretrained model for a specific downstream task by transferring the source model to a target domain. Recent works on vision-language model tuning include CLIP-Adapter~\cite{gao2021clip} that proposes to conduct fine-tuning with an extra bottleneck layer while freezing the pre-trained CLIP model. WiSE-FT~\cite{wortsman2022robust} resolves the distribution shift caused by fine-tuning and ensembles the weights of the original and fine-tuned model to increase robustness. Prior test-time adaption works~\cite{Wang2021TentFT,Wang2021OntargetA,Sun2020TestTimeTW} fine-tune on target data without information from the source data. Our approach leverages test-time adaptation for updating our meta-personalized model to user-specific data.

\noindent \textbf{Meta-Learning.}
We draw inspiration from meta-learning (``learning to learn'')~\cite{Finn2017ModelAgnosticMF,finn2019online,yu2020meta}, which enables models to quickly adapt to new tasks by learning on a diverse set of tasks. Hence, given only a handful of novel training examples, the model can be adapted to novel tasks. Our approach borrows the idea of meta-learning to meta-personalize a model by learning global category features from a large  video database. We then adapt the global features during test-time training from only few examples of user-specific instances to enable query-time retrieval.

%% file: sections/04_method.tex
\section{(Meta-)Personalization of Named Instances}
\label{sec:method}

Our goal is to learn representations of personal items in video that enable retrieval through natural language queries. 
To achieve this goal, we have to address the key challenge of adapting a model with one or few examples of a named instance.
To address this challenge, we propose a {\it meta-personalization} approach that learns to personalize given a large corpus of named instances mined from videos with transcriptions. We illustrate our approach in Figure~\ref{fig:overview}. 

In the first step (Figure~\ref{fig:overview} (left)), we mine automatically a collection $\mathcal{D}$ of named instances from videos with transcriptions for meta-personalization. We use this collection to train a proposed model $\mathcal{M}_{C,z}$ that includes global category features $C$ and instance-specific parameters $z$. The global category features $C$ are lightweight and shared across all instances. Given a natural language query $u$ and video $v$, the model returns a score $\mathcal{M}_{C,z}(u,v)$. During meta-personalization, given a training loss $\mathcal{L}$, we jointly optimize the loss over the global category features $C$ and  instance parameters $\mathcal{Z}$ for each named instance in the collection $\mathcal{D}$,
\begin{equation}\label{eq:metamodel}
(C_\mathcal{D}, \mathcal{Z_D}) \in \argmin_{(C,\mathcal{Z})} \sum_{z\in\mathcal{Z}} \mathcal{L}(C, z).
\end{equation}
Note that here the instance-specific parameters $\mathcal{Z}_\mathcal{D}$ learnt via~\eqref{eq:metamodel} 
 are discarded while the global category features $C_\mathcal{D}$ are kept as the meta-personalized part of the model. The global category features $C_\mathcal{D}$ capture information shared across instances relevant to the personalization task.

In the second step (Figure~\ref{fig:overview} (middle)), we are given a set of named video instances $\mathcal{P}$ (\eg, automatically mined from someone's  personal video library) and wish to perform test-time personalization of the model to this person's instances. 
Here each instance is represented by only one or few examples. In this step, we optimize the training loss $\mathcal{L}$ over the global category features $C$ and the set of instance parameters $\mathcal{Z}$ for all instances in the personal set $\mathcal{P}$ starting from the pre-trained global category features $C_\mathcal{D}$ and random $\mathcal{Z}$. We obtain the personalized model parameters $(C_{\mathcal{P}}, \mathcal{Z}_{\mathcal{P}})$ as
\begin{equation}\label{eq:testtime}
(C_{\mathcal{P}}, \mathcal{Z}_{\mathcal{P}}) \in \argmin_{(C, \mathcal{Z})}
\sum_{z\in\mathcal{Z}}
\mathcal{L}(C, z).
\end{equation}
We now keep both $C_{\mathcal{P}}$ and $\mathcal{Z}_{\mathcal{P}}$. 

In the final step (Figure~\ref{fig:overview} (right)), we perform retrieval over a potentially large dataset using the test-time personalized model $\mathcal{M}_{C_{\mathcal{P}},z}$ (where $ z\in \mathcal{Z}_\mathcal{P}$). We next describe how we automatically mine the named instances in video $\mathcal{D}$ for meta-personalization (Sec.~\ref{sec:mining}), and our full model $\mathcal{M}$ for query-time retrieval and loss $\mathcal{L}$ for meta-personalization and test-time personalization (Sec.~\ref{sec:learning}). 

\subsection{Automatic Mining of Named Instances in Video}
\label{sec:mining}
To identify personal instances without explicit supervision, we leverage a collection of videos from the web along with their corresponding time-aligned transcripts.
These transcripts can be automatically generated through speech-to-text models \cite{wang2020fairseq, whisper}.
We now describe a procedure to mine these data for a set of moments (\ie, a collection of video shots) depicting a referred personal instance in the transcript without the need for manual annotation.
We will use these moments for training a meta-personalization model (Sec.~\ref{sec:learning}).

\begin{figure}[t]
    \centering
    \includegraphics[width=\linewidth]{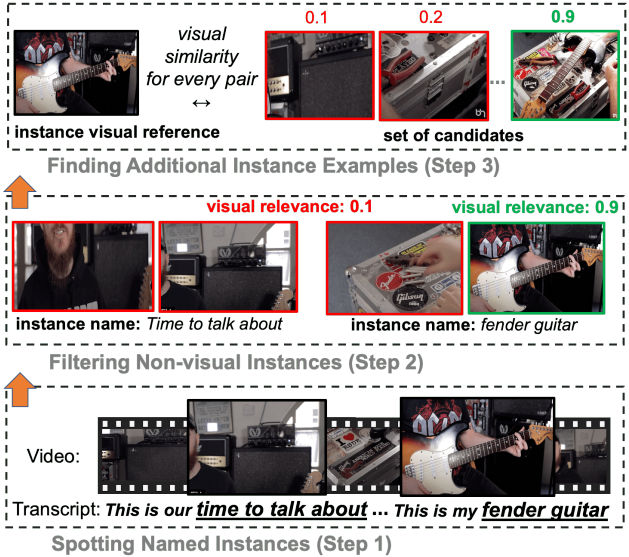}
    \vspace*{-8mm}
    \caption{\textbf{Automatic mining of named instances in video for meta-personalization.}
    Our automatic mining pipeline includes three steps (from bottom to top). {\bf Step 1} finds named instances via string-matching of possessive patterns in video transcripts. {\bf Step 2} filters non-visual instances using text-to-visual relevance between the instance name and the video shots neighboring the named instance. Finally, {\bf Step 3} retrieves additional shots with high visual similarity to the instance reference shot.
    }
    \label{fig:mining}
    \vspace*{-3mm}
\end{figure}

\noindent \textbf{Spotting Named Instances.}
Here our goal is to find moments where candidate personal instances are mentioned in videos. We do so by searching for possessive text patterns\footnote{\possesivepatterns} such as \texttt{"This is my $*$"} in a corpus of time-aligned video transcripts. This string-matching process outputs a list of candidate instance names $*$ associated with video timestamps $t^*$. In practice, we keep up to four words after a possessive text pattern is matched based on text-visual similarity (see appendix for details). That way, we can retrieve simple named instances such as \texttt{This is my dog} ($*=$\texttt{dog}) but also complex ones like  \texttt{This is my favorite CHANEL classic handbag} ($*=$\texttt{CHANEL classic handbag}). Note also that a single video might include multiple string matches; for instance, the example video illustrated in Figure~\ref{fig:mining} (Step 1) includes two matches: \texttt{"This is our time to talk about"} at time \texttt{1:30} and \texttt{"This is my fender guitar"} at time \texttt{3:25}. 

\noindent \textbf{Filtering Non-visual Instances.}
The previous spotting step only searches for potential instances using the transcript, yielding many string matches that are {\it non-visual}, \ie, the strings do not describe the visible content in the video. 
Here we aim to filter out these non-visual instances. 
We do so by computing the text-to-visual relevance between the instance name (\eg, \texttt{fender guitar}) and the neighboring shots around the time when the instance is mentioned. 
We add neighboring shots to cover cases where the named instances are shown just before or after they are mentioned.
Concretely, given a sequence of $m$ video shots $S=[s_1,\dots,s_m]$ automatically extracted with \cite{soucek2020transnetv2}, we find the shot $s_{t^*}$ that overlaps with $t^*$ (the time when the instance was mentioned). 
Next, we form a set of candidate visual references $S_{t^*}=[s_{t^*-1}, s_{t^*}, s_{t^*+1}]$ comprising a window of shots that are previous and subsequent to $s_{t^*}$. 
We then compute text-to-visual relevance scores using CLIP \cite{radford2021learning} encoders.
This encoding process yields $L_2$-normalized embeddings $f_l(*)$ for the named instance and $f_v(s_i)$ for each shot $s_i \in S_{t^*}$. We compute $f_v(s_i)$ by averaging the visual embeddings of all frames in the corresponding shot. 
Finally, we compute the cosine similarity between every $(f_l(*), f_v(s_i))$ pair and retain a visual reference shot $s^*$ if the highest cosine similarity is greater than $0.3$.
This filtering step outputs a cleaned set of named instances with a corresponding visual reference. Figure \ref{fig:mining} (Step 2) illustrates how we prune out non-visual matches such as \texttt{time to talk about}. In contrast, visual instances such as \texttt{fender guitar} are kept and matched with a visual reference.

\begin{figure}[t]
    \centering
    \includegraphics[width=\linewidth]{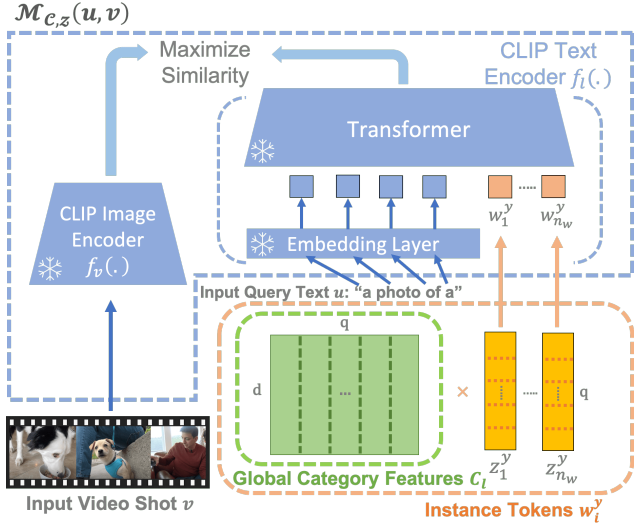}
    \vspace*{-8mm}
    \caption{\textbf{Model Overview.}
    Our model  $\mathcal{M}_{C,z}$ extends CLIP's language input vocabulary with $n_w$ novel instance-specific tokens $w_i^y=C_l z_i^y$, which we model as a linear combination of meta-personalized category features $C_l$ with weights $z_i^y$. 
    Note that the vision and language encoders are frozen during this process. 
    }
    \label{fig:model}
    \vspace{-3mm}
\end{figure}

\noindent \textbf{Finding Additional Instance Shots.}
Since frames from a single video shot provide only limited variability in the instance appearance and could thus limit the learning, we aim to recover other shots from the video where that instance appears.
We leverage CLIP's visual encoder to compute the visual similarity between the instance's reference shot $s^*$ and every shot $s_i \in S$. 
We extract an embedding $f_v(s^*)$ for the reference shot $s^*$ and an embedding $f_v(s_i)$ for each shot $s_i$. 
Similar to the non-visual filtering step, we average the CLIP embeddings of each frame belonging to a shot.
Then, we compute the cosine similarity between the embeddings for the reference shot and every candidate shot in the video. We keep the shots whose cosine similarity with the reference is greater than $0.9$.
Figure \ref{fig:mining} (Step 3) illustrates the output of this final step in the automatic mining pipeline. Our mining algorithm allow us to retrieve additional shots for the instance \texttt{fender guitar}. Now the instance examples not only include a clean close-up of the guitar, but also shots where the guitar is being played or held by its owner.

\begin{figure*}[t!]
\begin{center}
\includegraphics[width=0.85\textwidth]{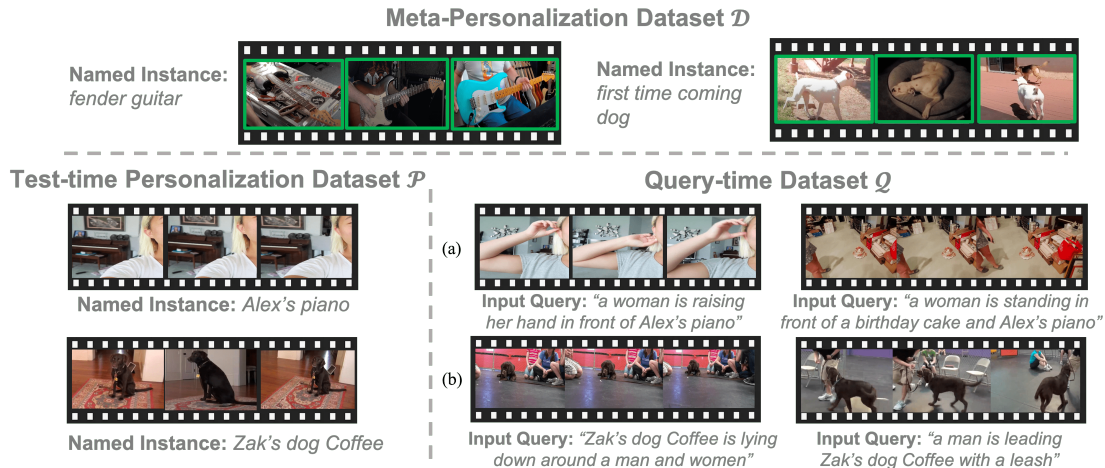}
\vspace*{-4mm}
\caption{\textbf{Examples from \datasetname \{ Meta-Personalization $\mathcal{D}$ (top) vs Test-time personalization $\mathcal{P}$ (bottom-left) vs Query-time $\mathcal{Q}$ (bottom-right)\} datasets. } In the Query-time dataset (bottom-right), we design a challenging video instance retrieval task. 
For example, in (a) the named instance (\ie, Alex's piano) is in the background and is barely visible and in (b) the background scenes in the query-time dataset (bottom-right) are completely different from the test-time personalization dataset (bottom-left) depicting the same named instance.}
\label{fig:thisismy_examples}
\vspace*{-8mm}
\end{center}
\end{figure*}

\subsection{Learning Personal Instance Representations}
\label{sec:learning}
The result of the mining procedure described above is a dataset consisting of a set of video shots $\mathcal{D}=\{s_1, \ldots, s_n\}$ and corresponding instance IDs $Y=\{y_1, \ldots, y_n\}$, where $y_i=y_j$ if $s_i$ and $s_j$ are video shots that are assumed to contain the same instance.
We now describe how we use these data to learn representations of the collected instances. 

Our approach is to leverage a large-scale pre-trained vision-language model (CLIP \cite{radford2021learning}) and augment its language encoder with a set of novel personal instance tokens.
Let $f_v(s)$ be the output of the visual encoder for shot $s$ (computed as the average over the frame embeddings) and let $f_l(u)$ be the output of the language encoder for a natural language input $u=[v_1, \ldots, v_m]$ of $m$ token embeddings, where $v_i \in \mathbb{R}^d$ denote learned token embeddings of the pre-training vocabulary (positional embeddings are included but omitted from the notation).
We propose to extend this vocabulary with novel personal instance tokens. 
Concretely, our approach introduces a set $w^y=\{w^y_{i}\}_{i=1}^{n_w}$ of $n_{w}$ new tokens that represent a personal instance $y\in Y$.

To learn these tokens and to perform personalized retrieval at test time, we construct natural language personalized queries as 
\begin{equation}
    \hat{u}^p = [p_1, \ldots, p_{k-1}, w^{y}_1, \ldots, w^{y}_{n_w}, p_{k+1}, \ldots, p_{m}],
\end{equation}
where $p_i$ are token embeddings of a chosen prompt $p=[p_1, \ldots, p_m]$.
During training, the prompt $p$ corresponds to a random template of the form \texttt{[An image of *]}, \texttt{[* can be seen in this photo]}, \texttt{[There is * in this image]}, etc., and $k$ denotes the position of the * placeholder for the instance tokens. 

\noindent \textbf{Instances as Combinations of Category Features.}
Since learning personal instance tokens from possibly very few examples runs the danger of overfitting (\eg, to nuisance features in the background), we propose to parameterize them as 
\begin{equation}
    w^y_{i}=C_l z^y_{i} \in \mathbb{R}^{d\times1},
\end{equation}
where $z^y_{i} \in \mathbb{R}^{q\times 1}$ is a vector of learnable weights specific to each instance and $C_l \in \mathbb{R}^{d \times q}$ is a matrix of learnable global category features, which are shared for all instances belonging to the same object category $l\in \mathcal{Y}$ (\eg, $\mathcal{Y} = \{ \text{car}, \text{person}, \text{dog}, \ldots\}$) and constitute the set $C_\mathcal{D}=\{C_l\}_{l\in \mathcal{Y}} $.
We illustrate the model in Figure~\ref{fig:model}.

We can think of the columns of $C_l$ as representing a set of shared category features and the final instance token $w^y_{i}$ as a linear combination of these features with weights $z^y_{i}$. 
Our aim is that only category-specific features are captured during training and irrelevant features (\eg, about the background) are discarded.
Intuitively, the columns of $C_l$ could correspond to attributes of an object category, \eg, if we were to learn ``car'' features, these could capture their color, brand, type, or age, to name a few. 
To identify which category matrix $C_l$ to use for an instance $y$, we rely on 0-shot classification using vision-language similarity between instance shots and a generic prompt for the category $l\in \mathcal{Y}$.

\noindent \textbf{Contrastive Personal Token Learning.}
We propose a contrastive learning objective to learn the personal instance tokens $w^y$ using a set of video shots containing the instance $y$.
To this end, let $\psi_i:=f_v(s_i)$ be an encoding of a video shot $s_i$, \ie, the average frame encoding of all frames in the shot, and let $\phi_i:=f_l(\hat{u}_i^p)$ denote the language encoding of a corresponding personalized query. 
We learn the novel tokens $w^y$ and shared category features $C_l$ by optimizing two contrastive objectives: a language-language contrastive objective $\mathcal{L}_{l}$ and a vision-language contrastive objective $\mathcal{L}_{vl}$.
The language-language objective is given by
\begin{equation}
\mathcal{L}_{l} = \sum_{i\in \mathcal{B}}  \sum_{j \neq i \in \mathcal{B}} -\mathds{1}\{y_i=y_j\} \log \left( \frac{d\left( \phi_i, \phi_j \right)}{ \sum_{k \neq i \in \mathcal{B}} d\left( \phi_i, \phi_k \right)} \right),
\end{equation}
where $d(a, b):= \exp \left(\frac{1}{\lambda}  \frac{ {a}^\intercal { b}}{\Vert {a} \Vert_2 \Vert {b} \Vert_2} \right)$
measures the similarity between the feature vectors $a$ and $b$, $\lambda=0.1$ is a temperature parameter, and $\mathcal{B}$ is a randomly sampled training mini-batch.
The vision-language objective is similarly defined as
\begin{equation}
\mathcal{L}_{vl} = \sum_{i,j \in \mathcal{B}} -\mathds{1}\{y_i=y_j\} \log \left( \frac{d\left( \phi_i, \psi_j \right)}{ \sum_{k \in \mathcal{N}} d\left( \phi_i, \psi_k \right)} \right),
\end{equation}
with the set of negative examples $\mathcal{N}$ comprising both other examples in the batch $\mathcal{B}$ and non-instance shots from the videos containing the named instances, \ie, shots that have low vision-language similarity. 
The loss is low when the encodings for video shots and personalized queries with the same instance ID are more similar than to other queries. Including non-instance segments as negatives can help discard non-instance features such as scene background. 

To further constrain the learning toward category-specific attributes, we include a loss that maximizes the similarity between a personal instance query and a generic category query. 
Concretely, let $c_l$ be a category query embedding for category $l$ (\eg, ``An image of a [car]'') to which instance $y$ belongs. 
We then include the following category-anchoring loss 
\begin{equation}
\mathcal{L}_{c} = - \sum_{i \in \mathcal{B}} \frac{ {c_l}^\intercal {\phi_i}}{\Vert {c_l} \Vert_2 \Vert {\phi_i} \Vert_2}.
\end{equation}

To summarize, our training loss $\mathcal{L}$ (see Equations~\ref{eq:metamodel} and \ref{eq:testtime}) for meta- and test-time personalization is given by 
\begin{equation}\label{eq:obj}
    \mathcal{L} = \mathcal{L}_l   + \mathcal{L}_{vl}  + \lambda_c \mathcal{L}_{c},
\end{equation}
where $\lambda_c=0.5$, controls the amount of category anchoring.
\\

%% file: sections/05_dataset.tex
\section{\datasetname Dataset}

Our \datasetname dataset comprises three subsets for meta-personalization, test-time personalization, and querying. We describe each subset next.

\noindent \textbf{Meta-Personalization Dataset $\mathcal{D}$.}
We gather this subset with the automatic mining algorithm introduced in Section \ref{sec:mining}. We leverage the Merlot Reserve dataset \cite{zellers2022merlot}, which contains more than 20 million videos from YouTube and their corresponding time-aligned transcripts. In practice, we start from a subset of $50K$ randomly sampled Merlot Reserve videos. We spot $6058$ named instances using various text possessive templates. Our visual filtering step removes $~52\%$ of instances, generating a total of $2908$ named instances with a visual reference. Finally, we mine additional samples for each instance, yielding a total of $49256$ instance samples. This subset includes a wide variety of visual concepts, ranging from common objects such as bikes to rare concepts such as toaster. While we design our mining pipeline to minimize noise, it is limited by CLIP's capability to distinguish between similar object instances. For instance, while we find several samples of the \texttt{fender guitar} instance, it includes other shots that do not correspond to the aforementioned guitar (Figure~\ref{fig:thisismy_examples} (top)). Nevertheless, we will show empirically that this subset is still useful for meta-personalization purposes (Section 5).

\noindent \textbf{Test-time Personalization Dataset $\mathcal{P}$.}
Our goal is to create a test dataset that recreates the scenario where a person wants to find when their personal instances appear in their video collection. We want to make this task close to the real scenario where a person records the same visual instance, \eg, their dog, across multiple places and situations. Our strategy is to emulate such a scenario by finding YouTube channels that frequently mention the same instance across multiple videos. While Merlot Reserve is large and diverse, only few channels are represented with more than one video in the dataset. Instead, we download all videos and automatic transcripts from the channels of $15$ popular YouTube bloggers. We then run our mining algorithm, but manually supervise the steps for visual filtering and finding additional instance samples. We first manually verify that each instance name and its visual reference are good matches. Then, we find additional sample shots across \textit{all videos in the channel} by ranking them according to their visual similarity to the instance reference shot. Finally, we review and label the top $1000$-scored shots for each named instance. The \textbf{appendix} includes a screenshot of the annotation tool. In total, this subset includes $15$ named instances with more than 686 labeled samples. Figure~\ref{fig:thisismy_examples} (bottom left) shows two example instances in our dataset.

\noindent \textbf{Query-time Dataset $\mathcal{Q}$.}
Our end goal is to retrieve named instances via natural language queries. We would like to be able to find videos when \texttt{<my dog biscuit> is grabbing a pink Frisbee.} To this end, we manually caption $30$ instance samples with descriptive queries. Thus, the Query-time dataset includes (manually captioned) video-caption pairs containing instances from the Test-time Personalization dataset. Figure~\ref{fig:thisismy_examples} (bottom right) shows manually curated captions for two instances in our dataset.

%% file: sections/06_results.tex
\section{Experiments}
\label{sec:exp}
Our experiments first ablate our model's contributions and loss design (Section~\ref{sec:ablations}), and evaluate our final model in personalized instance retrieval benchmarks (Section~\ref{sec:comp}).

\noindent \textbf{Evaluation Datasets and Metrics.} 
We evaluate our approach on two datasets: (i) our newly introduced \datasetname personal video instance retrieval benchmark, and (ii) the personalized image retrieval benchmark built on DeepFashion2 \cite{ge2019deepfashion2} proposed by \cite{cohen2022my}.
We evaluate retrieval performance in two settings: \\
\noindent \textbf{1) Generic Instance Retrieval:} In this case, we build queries for learned instances using a generic prompt, \ie, ``An image of *'', and measure retrieval performance using mean Average Precision (mAP) and Mean Reciprocal Rank (MRR).
In this setting, there are multiple correct matches.  \\
\noindent \textbf{2) Contextualized Instance Retrieval:} In this case, we use natural language queries that describe the personal instance in a specific context, \eg, using a scene description such as ``A photo of * lying on the beach.''. In this case, we assume there is only a single correct match, and we measure performance with MRR and Recall-at-5 (R@5). 

\noindent \textbf{Implementation Details.}
We use the ViT-B/16 version of CLIP in most of our experiments if not otherwise indicated.
All the learnable parameters $z_i$ and $C_l$ are randomly initialized from $\mathcal{N}(0, 0.1)$.
We set the number of category features to $q=512$ and the number of instance tokens to $n_w=1$ by default.
No data augmentation is used during training (the visual embeddings thus have to be computed just once). 
For meta-personalization of category features $C_l$, we randomly select 32 named instances from $\mathcal{D}$ that are 0-shot classified to each category $l$ and train for 20 epochs. 
This process is repeated 10 times, re-initializing $z_i$ each time while retaining $C_l$ from the previous run.  
Although updating the parameters of $C_l$ to a small set of instances $\mathcal{P}$ at test time is beneficial, there is a risk of overfitting when only a few or just a single instance of each category is provided.
To mitigate this, we include additional instances from $\mathcal{D}$ with the same categories as contained in $\mathcal{P}$ during test time personalization.
All our results are averaged over five runs, each with different random seeds, and we report the standard error for each experiment. 

\subsection{Ablations}\label{sec:ablations}

\begin{table}[t]
\centering
\caption{\textbf{Ablation Experiments.}
We verify our model and training objective design through ablations on \datasetname and DeepFashion2. 
We report personal instance retrieval performance in terms of mAP and MRR (higher is better). 
}\label{tab:ablations}
\footnotesize
\vspace*{-3mm}
\begin{tabular}{@{}l@{\hspace{1em}}c@{\hspace{1em}}c@{\hspace{1em}}c@{\hspace{1em}}c@{}}
\toprule
     &  \multicolumn{2}{c}{\textbf{\datasetname } } &  \multicolumn{2}{c}{\textbf{DeepFashion2}} \\ 

\textbf{Ablation}  & mAP  & MRR  & mAP  & MRR  \\ \midrule
a) w/o meta-pers.  & 54.1$\pm$1.3  & 83.1$\pm$2.1   &  35.2$\pm$0.5   &  55.2$\pm$1.9     \\
b) single $C$  & \underline{55.5}$\pm$0.4  & \underline{86.9}$\pm$0.5  &    44.1$\pm$0.5  &   64.7$\pm$1.5    \\
c) w/o $\mathcal{L}_l$  & 53.9$\pm$1.0  & 86.4$\pm$2.2  &    \textbf{47.3$\pm$}0.8  &   68.3$\pm$1.2    \\
d) w/o $\mathcal{L}_c$  & 48.9$\pm$1.0  & 78.6$\pm$2.4  &   \underline{47.1}$\pm$0.5  &   \underline{68.4}$\pm$0.8   \\
e) $\mathcal{N}=\mathcal{B}$  &  47.4$\pm$0.4 & 73.7$\pm$1.8  &   -   &   -   \\
f) w/o pre-trained $C$  & 53.0$\pm$0.7  & 85.1$\pm$3.0  &  44.3$\pm$0.8   &  67.1$\pm$1.0     \\
\midrule 
\rowcolor{LightGreen}
Ours   &   \textbf{56.4$\pm$0.6}  & \textbf{87.4$\pm$1.2}    & \textbf{47.3$\pm$0.7} & \textbf{69.9$\pm$0.8} \\
\bottomrule 
\end{tabular}
\label{tab.ablations}
\vspace{-3mm}
\end{table}

In Table~\ref{tab:ablations} we report results in the generic instance retrieval setting for the following ablations to illustrate the effect of our model and training objective design:\\
\noindent \textbf{a) w/o meta-personalization:}
In this case, we do not learn a meta-personalized $C$ and instead directly learn instance embeddings $w_i$.
This ablation demonstrates that learning global category attributes through meta-personalization improves generalization at test-time personalization.  
\\
\noindent \textbf{b) single $C$ (shared for all categories):} 
In this experiment, we learn only a single category matrix $C$, which is shared among all the categories. 
The results demonstrate that learning separate attributes per category is better than sharing global attributes among all categories.  
\\
\noindent \textbf{c) w/o $\mathcal{L}_l$:}
We explore the influence of the language-language contrastive loss $\mathcal{L}_l$. 
The benefits of  $\mathcal{L}_l$ are more pronounced in \datasetname, where instances belong to different categories.\\
\noindent \textbf{d) w/o $\mathcal{L}_c$:}
We analyze the contribution of the category-anchoring loss $\mathcal{L}_c$.
Including $\mathcal{L}_c$ is important on \datasetname, likely because it keeps the learning focused on the category and prevents it from capturing other scene features.
This is less the case on DeepFashion2, where images are very object-centric, and there is less background variety. 
\\
\noindent \textbf{e) $\mathcal{N}=\mathcal{B}$ (w/o non-instance segments):} We do not use non-instance segments $\mathcal{N}$ (\ie, segments of the same video but low vision-language similarity) as additional negatives in this ablation.  
Including $\mathcal{N}$ is important, as it can help prevent the instance embedding from capturing non-instance nuisance features from the scene. 
\\
\noindent \textbf{f) w/o pre-trained $C$:}
In this case, we randomly initialize $C$ instead of using a meta-personalized $C$.
Compared to a) the sharing of category attributes among fashion items in DeepFashion2 shows clear benefits. This is less the case for \datasetname, where instances belong to different categories.

\subsection{Personal Instance Retrieval Benchmarks}\label{sec:comp}

\noindent \textbf{Baselines.}
To demonstrate the effectiveness of our approach, we compare it to prior work \cite{cohen2022my} and the following strong baselines: \textbf{CLIP (visual)}, which represents each instance as the average visual embedding of the corresponding training examples, \textbf{CLIP (language)}, which represents each instance with a 0-shot prompt embedding of the corresponding category name, and \textbf{CLIP (V+L)}, which uses the average of the visual and language embedding. 

\begin{figure}[t]
    \centering
    \includegraphics[width=\linewidth]{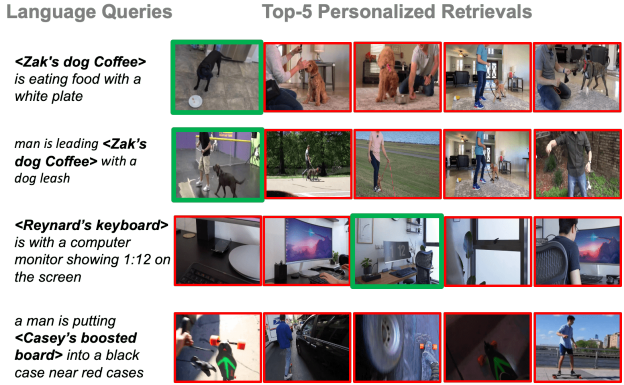}
    \vspace*{-7mm}
    \caption{\textbf{Contextualized \datasetname Retrievals.}
    We show personalized query-time retrievals for four \datasetname instances. 
    Search prompts are shown on the left and correct retrievals are highlighted in green.
    }
    \label{fig:retrievals}
    \vspace{-7mm}
\end{figure}

\noindent \textbf{Named Video Instance Retrieval on \datasetname.}
We evaluate our approach on the 15 named instances in our \datasetname test-time personalization dataset $\mathcal{P}$. 
To learn personal instance tokens, we train on video shots that occur in the video where the instance was named. 
All other video shots (belonging to other videos) are in the retrieval corpus for evaluation. 
Furthermore, we also include all other shots that do not contain the instance in the retrieval corpus as distractors. 
For contextualized retrieval we use the manually collected text captions from the query-time dataset $\mathcal{Q}$ as queries. 
We compare against the baselines and report contextualized retrieval performance on the left and generic instance retrieval performance on the right of Table~\ref{tab:tim_comp}.
Our model clearly outperforms the baselines in both settings. 
Interestingly, the visual and language baselines have opposing advantages in generic vs.\ contextualized retrieval, while our model performs well in both settings. 
Qualitative retrieval results are shown in Figures~\ref{fig:retrievals}.

\noindent \textbf{Fashion Item Retrieval on DeepFashion2.}
We follow the setup of \cite{cohen2022my} and consider DeepFashion2 \cite{ge2019deepfashion2} for personalized fashion item retrieval. 
The dataset in this setting consists of 653 training and 221 evaluation images across 50 instances (\ie, unique fashion items). 
While our focus is on video retrieval, our model and training objective can be used for image retrieval with minimal adjustments. 
We perform meta-personalization by pre-training on other non-instance-labeled images of DeepFashion2 (using only a single training image per instance). 
For a fair comparison with \cite{cohen2022my}, the instance tokens are learned by randomly sampling $k=5$ images per instance for training. 
We compare to results from \cite{cohen2022my} and the baselines in Table~\ref{tab:df_comp}.
Results using contextualized queries provided by \cite{cohen2022my} are on the left, and generic instance retrieval performance on the right. Note that \cite{cohen2022my} does not provide results for generic instance retrieval so we leave these two columns empty (-).
Our approach outperforms the baselines and prior work by a large margin.

\begin{table}[t]
\centering
\caption{\textbf{\datasetname Video Instance Retrieval Task.}
We report personal instance retrieval performance in language-specified contexts (\eg, ``* catching a pink frisbee'') on the left and generic instance retrieval (\eg, ``An image of *'') on the right.   
}\label{tab:tim_comp}
\vspace*{-3mm}
\footnotesize
\begin{tabular}{@{}l@{\hspace{1em}}c@{\hspace{1em}}c@{\hspace{1em}}c@{\hspace{1em}}c@{}}
\toprule
    &   \multicolumn{2}{c}{\textbf{Context. Retr.}} &  \multicolumn{2}{c}{\textbf{Generic Retr.}} \\ 

\textbf{Method}  & MRR  & R@5  & mAP  &  MRR  \\ \midrule
Random   & 0.0$\pm$0.0  &  0.0$\pm$0.0  &  1.1$\pm$0.2  & 2.2$\pm$1.3 \\
CLIP (language)   &  30.8$\pm$0.0  & 36.7$\pm$0.0 & 16.6$\pm$0.0  & 44.2$\pm$0.0    \\
CLIP (visual)   &  10.3$\pm$1.0 &  12.0$\pm$1.6  & 48.0$\pm$0.6   & 75.0$\pm$3.2  \\
CLIP (V+L)   & 20.9$\pm$1.6 &  23.3$\pm$2.1  &  51.7$\pm$0.4   & 81.9$\pm$0.0  \\
\midrule
\rowcolor{LightGreen}
Ours     & \textbf{42.0$\pm$1.3}  &  \textbf{50.7$\pm$1.8}   &  \textbf{56.4$\pm$0.6}  & \textbf{87.4$\pm$1.2} \\
\bottomrule 
\end{tabular}
\vspace*{-3mm}
\end{table}

\begin{table}[t]
\centering
\caption{\textbf{Fashion Item Retrieval on DeepFashion2.}
We evaluate our approach on the personalized instance retrieval task with contextualized queries from \cite{cohen2022my} (left) and generic instance retrieval (right).
Results with * use ViT-B/32 instead of ViT-B/16.
}\label{tab:df_comp}
\vspace*{-2mm}
\footnotesize
\begin{tabular}{@{}l@{\hspace{1em}}c@{\hspace{1em}}c@{\hspace{1em}}c@{\hspace{1em}}c@{}}
\toprule 
    &   \multicolumn{2}{c}{\textbf{Context. Retr.}} &  \multicolumn{2}{c}{\textbf{Generic Retr.}} \\ 

\textbf{Method}  & MRR  & R@5  & mAP  &  MRR  \\ \midrule
Random   &  2.9$\pm$0.2 &  1.8$\pm$0.5  &  4.7$\pm$0.5  & 9.5$\pm$2.0  \\
CLIP (language)    &   21.2$\pm$0.0 &  25.3$\pm$0.0 &  8.3$\pm$0.0  &  16.9$\pm$0.0    \\
CLIP (visual)   & 14.2$\pm$0.3  &  17.3$\pm$0.3  &   20.6$\pm$0.4  &  43.7$\pm$1.1  \\
CLIP (V+L)   & 20.8$\pm$0.8  &   25.4$\pm$1.7  &   20.5$\pm$0.5  &  42.8$\pm$1.2  \\
Adapter* \cite{cohen2022my}   &  5.9$\pm$0.7 &  -  &   -  &  -  \\
COLLIE* \cite{cohen2022my,skantze2022collie}   &  7.9$\pm$0.7 &  -  &   -  &  -  \\
PALAVRA* \cite{cohen2022my}    & 28.4$\pm$0.7  &  39.2$\pm$1.3  &   -  &  -  \\
\midrule
\rowcolor{LightGreen}
Ours*    &  34.4$\pm$0.7  & 45.2$\pm$1.1  &  40.0$\pm$1.0   &  69.3$\pm$1.8   \\
\rowcolor{LightGreen}
Ours     &  \textbf{38.4$\pm$0.4} & \textbf{51.4$\pm$0.4} &  \textbf{53.4$\pm$0.4}   &  \textbf{77.7$\pm$0.6}   \\
\bottomrule 
\end{tabular}
\vspace*{-3mm}
\end{table}

%% file: sections/07_conclusion.tex
\section{Conclusion}

We have introduced a meta-personalization approach for learning to retrieve named instances in video given a natural language query. We demonstrated the effectiveness of our approach on two datasets and showed that it outperforms strong baselines. 
Our effort is a step towards vision-language models trained over a large number of general and personal concepts.
Our approach opens up the possibility of personalized language-based video editing, summarization, and generation, \eg, identify the key instances in a collection of footage and create an edited story about the instances using natural language.

%% file: sections/appendix.tex
\appendix

\section*{\Large Appendix} 
In Section~\ref{sec:appendixMining}, we first provide additional details of the algorithm for the automatic mining of named instances in videos. 
Then in Section~\ref{sec:appendixDataset}, we give additional details about the process of collecting annotations for our dataset. 
Section~\ref{sec:appendixImplementation} provides additional implementation details of our approach and the evaluation metrics. 
Section~\ref{sec:posner} explores alternative approaches for mining instances, \ie, using Part-of-Speech (POS) and Named Entity Recognition (NER). 
Section~\ref{sec:limit} discusses some limitations of our approach. 
Finally, Section~\ref{sec:appendixResults} discusses additional qualitative results of personalized retrieval.

\section{Automatic Mining of Named Instances in Video}
\label{sec:appendixMining}

\noindent\textbf{Spotting Named Instances.}
We provide more details here of how we spot named instances (Section 3.1 of the main paper). We keep up to four words after a possessive text pattern is matched based on text-visual similarity. Given a sequence of words $[q_1,\dots,q_4]$ we extract embeddings $f_l([q_1]), f_l([q_1, q_2]),\dots,f_l([q_1,\dots,q_4])$ with CLIP’s text encoder. We then compute the cosine similarity with the visual reference embedding $f_v(s^*)$ and keep the longest sequence of words with cosine similarity greater than $0.3$. This strategy allows us to find relatively clean instance names. For instance, let us appendixose we string-match the candidate instance \texttt{this is my "dog waggy he is"}. Our approach would allow us to keep \texttt{dog waggy} as the instance name. We obtain the cleaned name given that additional words, \texttt{he is}, would yield a lower than $0.3$ text-visual similarity.
\\

\noindent \textbf{Filtering non-visual instances.} 
Regarding the filtering procedure for non-visual instances outlined in Section 3.1, we find that high visual-language similarity is observed when the candidate named instance features nouns or phrases that distinctly describe a visible object instance in the video. High visual-language similarity occurs for both general object categories (such as "my car") and more specific descriptions ("my 2018 Honda Civic"). 
Thresholds for spotting and filtering named instances were determined using cross-validation on a small curated validation set.

\begin{figure}[t]
    \centering
    \includegraphics[width=\linewidth]{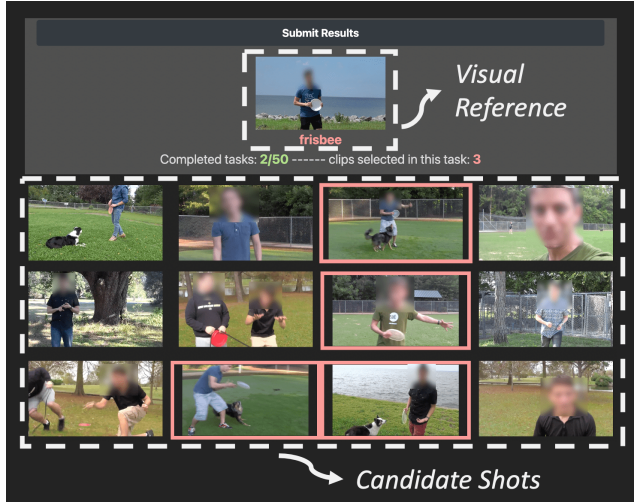}
    \caption{\textbf{Test-time Personalization Dataset annotation tool.} Our user interface contains two key components: (top) A video player that shows the visual reference of the target named instance. (bottom) A gallery of ``clickable'' candidate shots to be labeled as positives. The pink borders denote the selected positive samples, for the instance \texttt{"Zak's frisbee"}.
    }
    \label{fig:screenshot}
\end{figure}

\section{\datasetname Dataset}
\label{sec:appendixDataset}
\noindent\textbf{Test-time Personalization Dataset $\mathcal{P}$.}
We provide more details about our annotation tool (Section 4 of the main paper). Figure \ref{fig:screenshot} (this appendix) includes a screenshot of the annotation tool used to annotate the test-time personalization dataset. We implemented a simple user interface that shows the named instance (top) and a gallery of candidate shots (bottom). The interface auto-plays all candidate shots and allows the annotator to label the positive samples by clicking the video. Leveraging the interface, we are able to label the 1000 candidates for each instance in $20$ minutes. Therefore, we spent around five hours annotating the $15$ instances of the test-time personalization dataset.

\section{Additional Implementation Details}
\label{sec:appendixImplementation}

In all our experiments, we rely on the Adam optimizer \cite{kingma2014adam} with a weight decay set to $10^{-5}$. 
The learning rate follows a cosine annealing schedule \cite{loshchilov2016sgdr} with a maximum learning rate of $0.1$. Next, we describe the implementation details for the two datasets.
\\

\noindent \textbf{Baselines.}
In the CLIP (language) baselines, we pass only the manually labeled object category (\eg, "dog") to the text encoder, which prevents confusion caused by queries containing names of specific instances such as \texttt{"Zak's dog Coffee"} and \texttt{"My dog Biscuit"}.\\

\noindent \textbf{DeepFashion2 Experiments.}
Each test-time training is performed for 40 epochs with a batch size of 512. 
In this setting, learning the 50 instance tokens takes about 10 minutes in total on a single GPU.
We perform 10 rounds of meta-personalization for the pre-training of category features $C$, each round consisting of $32$ pseudo instances per category (only a single training image per instance is available). 
Each training round consists of 10 epochs, and we use a batch size of 512.
We identify 14 categories for DeepFashion2\footnote{List of DeepFashion2 categories:
\texttt{['long sleeve dress', 'long sleeve top and skirt', 'long sleeve top and trousers', 'long sleeve top and vest dress', 'short sleeve top and shorts', 'short sleeve top and skirt', 'short sleeve top and sling dress', 'short sleeve top and trousers', 'shorts and vest', 'skirt and sling', 'skirt and vest', 'sling dress', 'trousers and vest', 'vest dress']}}.
Our CLIP (language) baseline uses these categories in place of the learned instance tokens for retrieval. 
\\

\begin{table*}[t]
\centering
\caption{
\textbf{Alternatives for mining instances.}
We annotate $100$ candidate instances for different mining approaches, including, Part-of-Speech (POS) and Named Entity Recognition (NER). We report the number of true named instances and the precision for each method. Possessives w/ NER filter denotes our mining approach combined with a filter that discards candidate instances without recognized entities. Combining possessives with POS yields much lower precision, thus not included in this table.
}\label{tab:mining_ablation}

\small
\begin{tabular}{@{}l@{\hspace{1em}}c@{\hspace{1em}}c@{\hspace{1em}}c@{\hspace{1em}}c@{}}
\toprule
    &   \multicolumn{2}{c}{\textbf{w/o visual filter}} &  \multicolumn{2}{c}{\textbf{with visual filter}} \\ 

\textbf{Method}  & $\#$ named instances  & Precision  & $\#$ named instances & Precision  \\ \midrule
POS (nouns) & 19 &  19.0\%  &  15  & 36.6\% \\
NER &  21  & 21.0\% & 17  & 38.3\%    \\
\midrule

Possessives (ours)& 58  &  58.0\%   &  \textbf{46}  & 63.1\% \\
\rowcolor{LightGreen}
Possessives w/ NER filter & 48  &  64.0\%   &  39  & \textbf{70.5\%} \\
\bottomrule 
\end{tabular}

\end{table*}

\noindent \textbf{\datasetname Experiments.}
Each test-time training is performed for 40 epochs with a batch size of 16, which takes less than two minutes on a single T4 GPU.
We use 512 randomly chosen distractor shots at each training iteration during test-time personalization.
Meta-personalization consists of 10 rounds of training, with each round lasting for 20 epochs. We use a batch size of 512 and do not include any distractor shots.
\\

\noindent \textbf{Evaluation Metrics.}
For completeness, we provide definitions for the retrieval metrics used in our experiments. 
Let $R_{ij}$ be indicators of whether the retrieved video shot for query $i$  at rank $j$ is a correct match, \ie, $R_{ij}=1$ if the $j$-th shot retrieved for query $i$ is showing the correct instance (in the right context), and $R_{ij}=0$ otherwise.
Let further $\operatorname{rank}_i=\min \{j |R_{ij}=1\}$.
We then have 
\begin{equation}
   \operatorname{MRR} = \frac{1}{N} \sum_{i=1}^N \frac{1}{\operatorname{rank}_i}, 
\end{equation}
\begin{equation}
    \operatorname{R@K} = \frac{1}{N} \sum_{i=1}^N \mathds{1}\{\operatorname{rank}_i\leq K\} , 
\end{equation}
and 
\begin{equation}
    \operatorname{mAP} = \frac{1}{N} \sum_{i=1}^N \sum_{k=1}^K \frac{R_{ik}}{n_i} \operatorname{P}_{ik}   , 
\end{equation}
where $\operatorname{P}_{ik}=\frac{1}{k} \sum_{j=1}^k R_{ij}$ is precision-at-$k$ for the $i$-th query, $n_i=\sum_{j=1}^K R_{ij}$ is the number of relevant shots for query $i$, $N$ is the number of queries and $K$ the number of shots in the retrieval dataset.
\\

\noindent \textbf{Hyper-parameters.}
We use COCO \cite{lin2014microsoft} classes as the object categories $l\in \mathcal{Y}$, since they represent common
and general objects.
We set the temperature parameter $\lambda=0.1$ to a standard default value
used in contrastive losses (\eg, SimCLR \cite{chen2020simple}).
We choose $\lambda_c=0.5$ through cross-validation and find that our model's performance is robust with respect to different values of $\lambda_c$.

\section{Alternatives for Mining Instances}
\label{sec:posner}
We explore alternative approaches for spotting named instances (Section 3.1 of the main paper). 
In our approach, we set the list of possessive text patterns for mining empirically. 
After annotating a small sample, we observed that those patterns yield a larger number and more precise set of named instances compared to other alternatives such as Part-of-Speech (POS) and Named Entity Recognition (NER) (see Table \ref{tab:mining_ablation} in this appendix). 
Interestingly, combining our approach for possessive text pattern matching with an additional NER filter can improve the precision of the mined instances.
For simplicity, we do not apply the NER filter for our final collected dataset as described in the main paper.

\section{Discussion}
\label{sec:limit}
We discuss potential limitations of our work and highlight key differences to prior work \cite{cohen2022my}.\\

\noindent \textbf{Handling of multiple subjects featured in a video.}
During data collection, we did not regulate the number of subjects featured in each clip; as a result, there could be instances where multiple subjects are presented. These multiple subjects may affect the precision of our mining approach, particularly if the subjects belong to the same visual category. Nevertheless, we have observed that specific objects are usually visually conspicuous when mentioned. For example, when the speaker mentions \texttt{This is my dog <Fido>}, a close-up or zoom-in shot of the dog are typically shown. In future work, the narration's contextual information (\eg, \texttt{<Fido> eating}) could be utilized to differentiate instances with several subjects.\\

\noindent \textbf{Size of the \datasetname test-time personalization dataset.}
Our \datasetname dataset contains a modest number of instances; however, the search space is very large (around 50K shots, including distractors). We also acknowledge that creating this dataset posed a considerable challenge, as it required identifying instances across numerous videos and annotating each shot per video. 
\\
\noindent \textbf{Differences to PALAVRA \cite{cohen2022my}.}
Our work distinguishes itself from \cite{cohen2022my} in three crucial aspects:
\begin{enumerate}
\item \textbf{Model:} \cite{cohen2022my} models instance tokens independently, whereas our method represents them as a weighted sum of shared category features learned through meta-personalization. As demonstrated in ablation (a) of Table 1 in the main paper, our design improves the model's generalization capabilities.

\item \textbf{Training Data:} Unlike \cite{cohen2022my}, which requires a set of labeled examples per instance, we propose a method to mine training examples from narrated videos.

\item \textbf{Training Objective:} Our method proposes a contrastive training objective for test-time personalization, whereas \cite{cohen2022my} requires additional networks (see set encoder in \cite{cohen2022my})

\end{enumerate}

\noindent \textbf{\datasetname vs. YTVOS \cite{xu2018youtube} for personal video instance retrieval.}
In contrast to \cite{cohen2022my}, which used YTVOS by taking query and retrieval frames from the same video, we explore a more challenging scenario where query and retrieval shots are from different videos, showing the instance in different contexts. 

\section{Qualitative Results for Contextualized Instance Retrieval}
\label{sec:appendixResults}
\begin{figure*}[t!]
\begin{center}
\includegraphics[width=\textwidth]{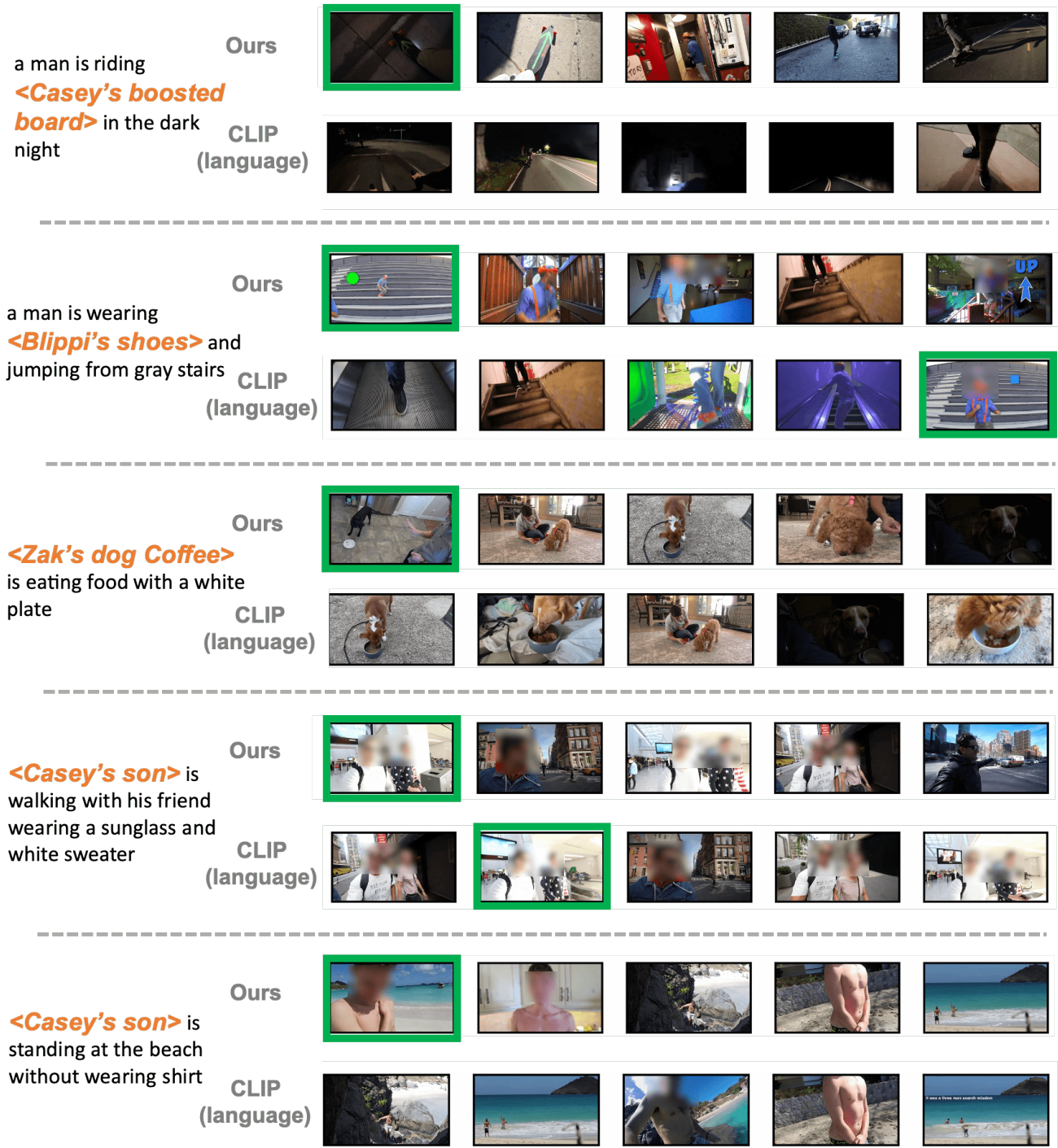}
\caption{\textbf{Qualitative Retrieval Comparison to the CLIP (language) Baseline.} 
While the baseline is able to accurately match the features in the scene that match the described context, it fails to retrieve the correct instance. In contrast, our personalized VLM successfully matches both context and personalized instance. Search prompts are shown on the left and correct retrievals are highlighted in green.
}
\vspace{-3mm}
\label{fig:appendix-retrieval-comp}
\end{center}
\end{figure*}

Figure~\ref{fig:appendix-retrieval-comp} provides additional qualitative results for the contextualized instance retrieval task on the \datasetname dataset. It compares the Top-5 retrievals of our approach and the 
CLIP (language) baseline given a language query. Both methods can successfully retrieve shots that match the generic context of the query, \eg, \texttt{eating food with a white plate} (third row). However, the baseline fails at retrieving the correct personalized instance, \eg, \texttt{"Zak's dog Coffee"} (third row). In the example of \texttt{Casey’s son is standing at the beach without wearing shirt} (last row), the CLIP (language) baseline fails to find the personalized instance since it does not have a representation for the instance \texttt{"Casey’s son"}. Contrarily, our model finds the correct named instance in the right context. This result is due to our model expanding the input space of the VLM by personalizing a representation of the learned instance while maintaining the general abilities of the underlying VLM.

\begin{figure*}[t!]
\begin{center}
\includegraphics[width=\textwidth]{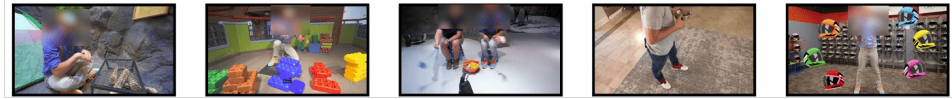}
\caption{\textbf{Qualitative examples of retrieval using our personalization approach.} 
Top: Successful examples where the correct instance is retrieved within the top five retrieved shots. Bottom:  Examples of failures where the correct instances is not retrieved within the top five retrieved shots. 
}
\vspace{-3mm}
\label{fig:appendix-retrieval}
\end{center}
\end{figure*}

Figure~\ref{fig:appendix-retrieval} shows successful and failure cases of our model on the contextualized instance retrieval task. (top) We observe that our method correctly retrieves the personalized instances even in challenging scenarios. For instance, it can retrieve small instances such as \texttt{"Casey's boosted board"} and \texttt{"Blippi's shows"} for different context queries. By keeping the VLM frozen, our method preserves the original VLM's capabilities to match natural language queries to the candidate set of shots. (bottom) While our method significantly improves the state-of-the-art in personalized retrieval, we observe some common failure cases. One typical example is discriminating between instances that are too similar. For instance, \texttt{"Sherry's road bike"} is confused with another black bike. Our method is also limited by the VLM's capabilities to understand actions such as \texttt{grabbing Sherry's road bike}. The former failure case could be addressed by leveraging additional cues from the transcript and the latter by leveraging progress on motion-aware VLMs.

